\documentclass[letterpaper, 10 pt, journal, twoside]{IEEEtran}  % Comment this line out if you need a4paper

\IEEEoverridecommandlockouts                              % This command is only needed if 
\usepackage{graphics} % for pdf, bitmapped graphics files
\usepackage{epsfig} % for postscript graphics files
\usepackage{amsmath} % assumes amsmath package installed
\usepackage{amssymb}  % assumes amsmath package installed
\usepackage{hyperref}
\usepackage{xcolor}
\usepackage{tikz}
\usetikzlibrary{calc}
\usetikzlibrary{arrows}
\usepackage{pgfplots}
\usepackage{siunitx}
\newcommand{\argmin}{\operatornamewithlimits{arg\ min}}

\pgfplotsset{compat=1.18}

\title{Robust Self-Tuning Data Association for Geo-Referencing Using Lane Markings
}

\author{Miguel Ángel Muñoz-Bañón$^{1}$,  Jan-Hendrik Pauls$^{2}$,  Haohao Hu$^{2}$, Christoph Stiller$^{2}$, \\ Francisco A. Candelas$^{1}$, and Fernando Torres$^{1}$% <-this % stops a space

\thanks{Manuscript received: July, 27, 2022; Revised September, 23, 2022; Accepted October, 17, 2022.}%Use only for final RAL version
\thanks{This paper was recommended for publication by Editor S. Behnke upon evaluation of the Associate Editor and Reviewers' comments.}
\thanks{This work has been supported by the regional Valencian Community Government and the European Regional Development Fund (ERDF) through the project PROMETEO/2021/075 and the grants ACIF/2019/088 and BEFPI/2021/069.}% <-this % stops a space
\thanks{$^{1}$Authors are with the Group of Automation, Robotics and Computer Vision (AUROVA), University of Alicante, San Vicente del Raspeig S/N, Alicante, Spain.
        {\tt\small miguelangel.munoz@ua.es}}%
\thanks{$^{2}$Authors are with Institute of Measurement and Control Systems, Karlsruhe Institute of Technology, Karlsruhe, Germany.}%
\thanks{Digital Object Identifier (DOI): 10.1109/LRA.2022.3216991}
}

\begin{document}

\maketitle
%\thispagestyle{empty}
%\pagestyle{empty}

%%%%%%%%%%%%%%%%%%%%%%%%%%%%%%%%%%%%%%%%%%%%%%%%%%%%%%%%%%%%%%%%%%%%%%%%%%%%%%%%
\begin{abstract}
Localization in aerial imagery-based maps offers many advantages, such as global consistency, geo-referenced maps, and the availability of publicly accessible data. However, the landmarks that can be observed from both aerial imagery and on-board sensors is limited. This leads to ambiguities or aliasing during the data association.

Building upon a highly informative representation (that allows efficient data association), this paper presents a complete pipeline for resolving these ambiguities. Its core is a robust self-tuning data association that adapts the search area depending on a pseudo-entropy of the measurements. Additionally, to smooth the final result, we adjust the information matrix for the associated data as a function of the relative transform produced by the data association process.

We evaluate our method on real data from urban and rural scenarios around the city of Karlsruhe in Germany. We compare state-of-the-art outlier mitigation methods with our self-tuning approach, demonstrating a considerable improvement, especially for outer-urban scenarios.

%Localization using aerial imagery-based maps, so-called geo-referencing, provides advantages such as global consistency, independence from the map construction process, and the amount of publicly accessible data. However, the limitation of landmark variety observable simultaneously from sensors and aerial imagery usually produce aliasing problems in outer-urban scenarios. This paper presents a complete geo-referencing pipeline using lane markings as landmarks. To address the outliers problems derived from aliasing, we performed a robust implementation providing self-tuning capabilities to our data association process by adapting the search area depending on the entropy in the measurements. The data representation used for entropy quantification and data association used for self-tuning are presented in previous work. Additionally, to smooth the final result, we adjust the information matrix for the associated data as a function of the relative transform produced by the data association process.

%We evaluate our method in urban and road scenarios with large aliasing risk areas from the city of Karlsruhe (Germany) by using our experimental vehicle BerthaOne. We compare state-of-the-art outlier mitigation methods with our self-tuning approach, demonstrating a considerable improvement, especially for the outer-urban scenarios. 

\end{abstract}


%%%%%%%%%%%%%%%%%%%%%%%%%%%%%%%%%%%%%%%%%%%%%%%%%%%%%%%%%%%%%%%%%%%%%%%%%%%%%%%%
\section{INTRODUCTION}

\IEEEPARstart{A}{utonomous} driving has become a major research topic over recent years. Systems such as self-driving cars depend strongly on their localization capabilities. One of the most comprehensive localization approaches is Simultaneous Localization And Mapping (SLAM) \cite{cadena2016past}, where a model of the environment (the map) is constructed while, at the same time, the vehicle's state is estimated.

In some applications, the map is assumed to be already known from dedicated mapping drives, third-party map suppliers, or the extraction from aerial imagery. While maps created from mobile vehicles offer high local accuracy, global consistency and geo-referencing are non-trivial issues when GNSS reception is impaired due to shadowing and multipath effects. It is worth noting that even when no overhead obstructions allow aerial imagery, the shadowing, and multipath effects could occur due to buildings, hills, or trees. Additionally, there are other inconsistencies with GNSS, e.g., the positioning at a fixed point can vary during the day depending on the position of the satellites. In contrast, maps that are extracted from aerial imagery do not have this problems as the aerial imagery is already geo-referenced and, hence, globally consistent. The localization in geo-referenced maps can be applied as an online localization system in a self-driving car, as an offline process to introduce global consistency to built maps, or even to merge maps created in different experimental sessions \cite{hu2019accurate}.

%In some applications, the map is available \cite{pauls2020monocular}, usually previously built by SLAM or even by LiDAR Odometry And Mapping (LOAM) \cite{zhang2014loam}. Then, the localization system consists of estimating the pose on that map. Localization and SLAM usually provide high local accuracy, but they are globally inconsistent due to drifts suffered from accumulated errors in constructing the environment's model. Another localization approach that avoids global inconsistencies is the so-called geo-referencing, where the environment information can come from aerial imagery, OpenStreetMaps (OSM) \cite{haklay2008openstreetmap}, or even by hand labeling. This localization can be applied as an online localization system in a self-driving car, as an offline process to introduce global consistency to built maps, or even to merge maps created in different experimental sessions \cite{hu2019accurate}.

\begin{figure}[t]
\centering
\includegraphics[width=200pt]{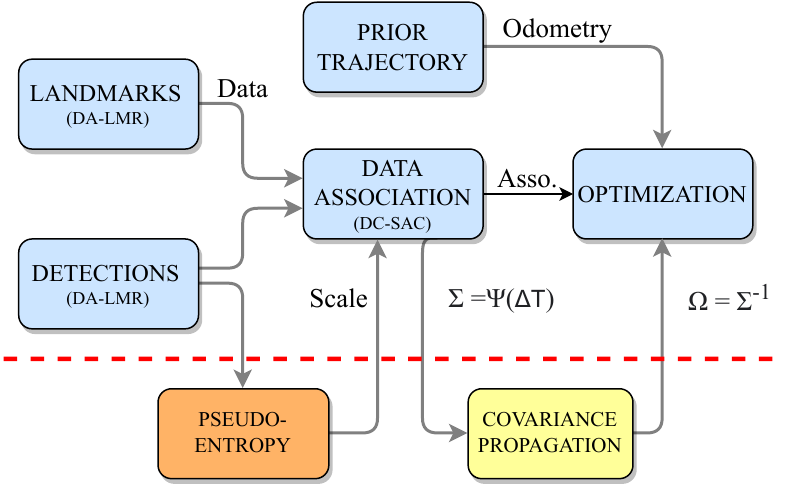}
\caption{The complete pipeline of our self-tuning geo-referencing approach. }
\label{fig:overview}
\end{figure}

%The blue blocks indicate the standard localization architecture, and the orange block calculates the data information used to tune the data association method. Both data representation and association are previously presented in \cite{munoz2022lmr}. The yellow block represents the covariance propagation to associations depending on the relative transformation of the DC-SAC results.

In theory, geo-referencing could be classified in the category of localization methods with a previously built map. However, it has particularities that can differentiate it from the other localization approaches in practice. For example, it has special requirements such as a global frame referenced prior, like GNSS, or an outdoor environment that is visible from planes or satellites. Also, as an environmental requirement, the landmarks selected to perform the localization must be observable for both satellite systems and on-board sensors. Such requirements cause sparsity or aliasing problems to the landmarks that meet them.

We consider lane markings an appropriate kind of observable landmarks for geo-referencing approaches. Buildings outlines are also observable for the both spaces of geo-referencing strategies, but we believe them to be highly restricted to urban areas. However, while lane markings usually offer sufficient information in lateral direction, they are often ambiguous in the direction of travel, leading to aliasing\footnote{For example, consider a vehicle circulating on a straight road that detects lane markings that could be two parallel lines (road edges). Consequently, we suffer from ambiguities if we try to associate such parallel lines on a map where the lane is also represented as lines more extended than the detected ones.}. This aliasing effect produces an unfavorable scenario where the data association suffers from many outliers. In other words, we can see an aliasing scenario equivalent to an high outlier regime. Most of the works in geo-referencing usually do not pay attention to aliasing and outliers problems because they are typically constrained to urban environments. In this work, we generalize to perform geo-referencing for both urban and rural scenarios. For that reason, we need to focus on avoiding the problem of aliasing and outliers.

For the past years, there has been a wide variety of works that aims to perform outlier mitigation in localization by using different approaches, such as convex relaxation~\cite{carlone2018convex} or covariance scaling \cite{agarwal2013robust}. Those approaches focus on robust performance of the objective function in the inference of graph-based probabilistic models. In contrast, automatic parameter tuning is another strategy to tune the back-end process depending on the low-level (the front-end) data information \cite{cadena2016past}. 

%We base our research on the last-mentioned approach. But, in our case, to provide improved data to the back-end, we self-tune the data association, that is, in the front-end. 

In this paper, we present a complete geo-referencing pipeline using lane markings as landmarks (Fig. \ref{fig:overview}). We base our system on robust performance by self-tunning capabilities to avoid the problem of aliasing and hence outliers derived from the use of lane markings. Given our delta-angle lane marking representation DA-LMR \cite{munoz2022lmr}, we quantify the information of the data in terms of the curvature of the lane marking for each data association execution. Then, by using our distance-compatible sample consensus DC-SAC~\cite{munoz2022lmr} data association, we can tune the search space depending on the information content of the measurements. In this way, when there is rich information like in intersections, DC-SAC can successfully search a wide space, making it robust to a bad prior. At the same time, in case of aliasing or poor information due to straight, solid markings, the data association is defaulting to a nearest neighbor search. Additionally, as DC-SAC is a pose estimation method, we adjust the covariance of the association depending on the pose results of the data association process. This kind of covariance adjustment smooth the evolution of the trajectory.

Our work contributions can be summarized as follows:

\begin{itemize}
    \item A complete pipeline for geo-referencing using lane markings to generalize the localization for both rural and urban environments by avoiding the aliasing effect.  
    \item Self-tuning data association where the search space depends on the information quantified by DA-LMR~\cite{munoz2022lmr}.
    \item Covariance adjustment that depends on the relative transformation resulting from the data association.
\end{itemize}

The rest of the paper is organized as follows: In Section \ref{sec:related_work}, we review the related works in geo-referencing and localization with outlier mitigation. In Section \ref{sec:optimization}, we describe the proposed model for the optimization problem. Then, the main contributions, self-tuning data association, and covariance adjustment are explained in Section \ref{sec:self_tuning} and Section \ref{sec:covariance}, respectively. Next, in Section \ref{sec:evaluation}, we present the evaluation results. Finally, Section \ref{sec:conclusions} exposes the main conclusions derived from this work and possible future works.

%%%%%%%%%%%%%%%%%%%%%%%%%%%%%%%%%%%%%%%%%%%%%%%%%%%%%%%%%%%%%%%%%%%%%%%%%%%%%%%%
\section{RELATED WORK}
\label{sec:related_work}

%In this section, we briefly review existing geo-referencing approaches. Additionally, we review localization approaches with outlier mitigation.

\subsection{Geo-referencing}

As we previously mentioned, the type of landmarks selected to perform geo-referencing should be observable from both aerial images and on-board sensors. Building walls are commonly used in a wide variety of works. In \cite{cho2022openstreetmap}, the authors obtain information about buildings from OSM and detect walls by using LiDAR sensors. In other work~\cite{roh2017aerial}, buildings are also used as landmarks, but in this case, the authors obtain the information directly by detection in aerial geo-referenced images. In \cite{yan2019global}, the authors use semantic descriptors performed also using buildings information. In such cases, buildings landmarks carry the dependency on navigating within urban environments. 

%In some works \cite{suger2017global}, the strategy of matching the vehicle's trajectory with lanes obtained from OSM or aerial imagery is used. Such approaches are not dependent on the urban environment in theory, but in practice, the achievable accuracy using lanes is considerably lower than when using individual lane markings. This holds particularly on outer-urban roads.

%Other approaches are based on learning methods \cite{fu2020lidar}, where a place is recognized from both aerial imagery and on-board sensors. A similar approach is the 

Other approaches are based on the so-called cross-view localization \cite{tian2017cross}, where local images are matched with aerial imagery. In \cite{downes2022city}, the authors implement Wide-Area Geo-localization (WAG) by combining a neural network with a particle filter for geo-localization in large aerial images. Such strategies usually suffer from feature sparsity in the rural areas and depend strongly on rich-information environments, such as urban. 

In \cite{hu2019accurate}, the authors use lane markings and poles as landmarks for geo-referencing, allowing localization in rural environments. However, we consider poles hard to detect from aerial imagery, so we choose only lane markings for this work. In \cite{hu2019accurate}, geo-referencing is performed to achieve global consistency of maps that were created using a SLAM algorithm with on-board sensor data.

All discussed geo-referencing works either cannot achieve comparable accuracy compared to using landmarks like lane markings, suffer detection challenges and/or suffer from aliasing. In contrast, we combine easily detectable landmarks that allow highly accurate geo-referencing with a robust solution for the aliasing problem that such landmarks suffer from. An alternative to the explicit handling of aliasing issues are localization methods with outlier mitigation.

%The cited geo-referencing works do not contemplate the problem of aliasing in the roads environment that produces outliers scenarios. For that reason, we also review localization approaches in such scenarios.

\subsection{Localization with outlier mitigation}

The localization problem can be divided into two main components: the \textit{front-end} and the \textit{back-end} \cite{cadena2016past}. The front-end involves landmarks detection, data representation, data association, etc. At the same time, the back-end infers the abstracted information provided by the low-level layer (the front-end). While the problem of aliasing and outliers in the data association process occurs at the front-end level, it is common to prevent its consequences on the inference, i.e., at the back-end layer. Traditionally, outlier mitigation in localization relied on robust M-estimators \cite{huber1996robust}, such as in \cite{kaess2012isam2}, where the authors use the Huber loss. Additionally to Huber loss, in \cite{agarwal2013robust}, the authors use a robust function that dynamically scales the covariance to reduce the influence of measurements with significant errors. An additional experimental analysis of the covariance scaling method is available in \cite{agarwal2014experimental}. A popular strategy to deal with outliers is the convex relaxation of the objective function in the optimization process, where \cite{carlone2014selecting,carlone2018convex} are seminal works. Instead of mitigating the effect of the outliers with a convex relaxation, Instead of mitigating the effect of the outliers with a convex relaxation, \cite{yang2020graduated} uses graduated non-convexity, whereas \cite{lusk2021clipper} and \cite{shi2021robin} are both graph-theoretic frameworks (\cite{lusk2021clipper} solves a continuous relaxation of the NP hard problem) and \cite{sun2021ransic} is inspired by these methods.

Previously mentioned works are focused on tuning the objective functions in the inference of graph-based probabilistic models. In contrast, other approaches aim at the problem in the whole back-end structure. For instance, in \cite{pfingsthorn2016generalized}, the authors model ambiguous measurements using hyperedges and a multimodal mixture of Gaussian constraints. Continuous and discrete graphical models are mixed in \cite{lajoie2019modeling} to avoid perceptual aliasing. In \cite{bloesch2017two}, the authors deal with a two-state implicit filter to perform outlier rejection.

Another robust performance to deal with outliers is the self-tuning strategy. Here, the front-end's data information can usually tune the objective function \cite{agamennoni2015self,santos2020mobile} or even adjusts the graphical model \cite{soldi2019self}. In \cite{lu2022slam}, the authors use automatic covariance tuning to estimate the pose of the environment's objects during SLAM process.

Dealing with outlier mitigation in the back-end is essential because the presence of outliers is expected, and the high-level layer must be ready. However, we focus our research on mitigating the outliers in the front-end to pass the information as cleanly as possible to the back-end. Concretely in the data association process where we perform self-tuning capabilities.

%Other previous works deal with outliers in a distance-compatible-based data association \cite{lusk2021clipper,sun2021ransic}, but both are designed in the context of point cloud registration, and their implementation requires data more geometrically descriptive than our DC-SAC. 

In the next section, we briefly introduce the graphical model defined for the back-end of the proposed geo-referencing.

\begin{figure}[t]
\centering
\includegraphics[width=200pt]{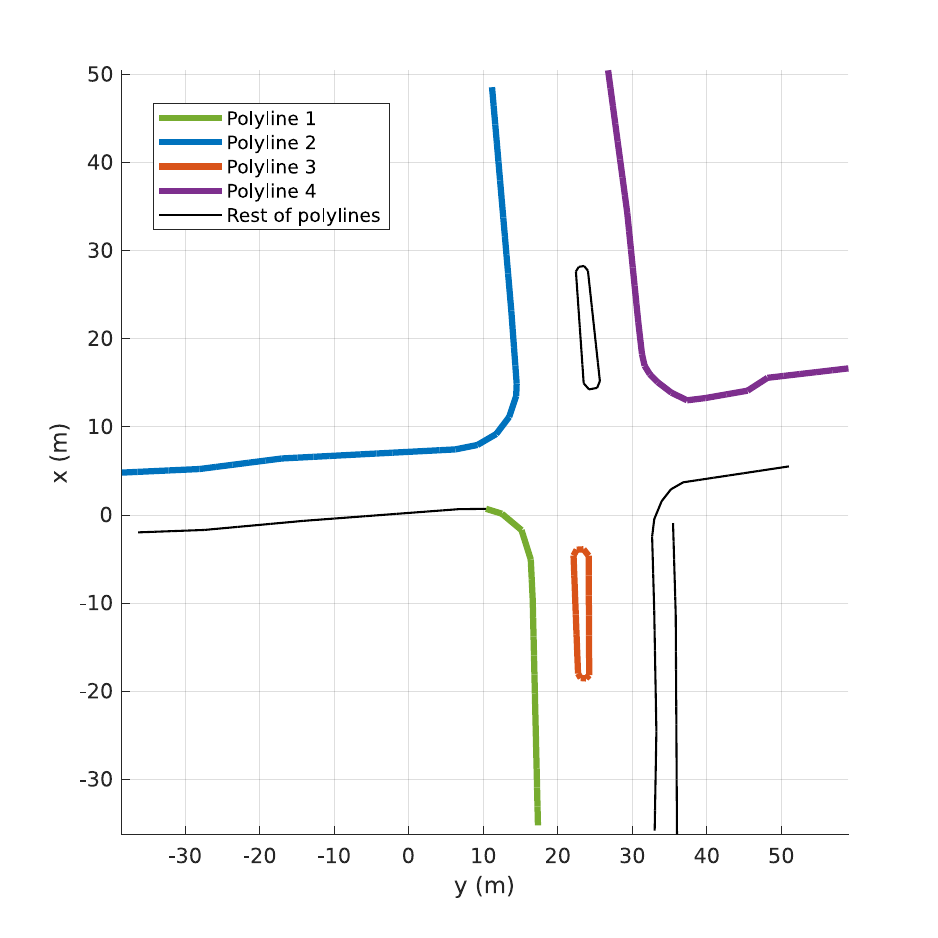}
\caption{Example of 2D projected polylines for a set of detections $\mathcal{D}_i$. We mark in colors four polylines that we represent as 1D signals in Fig. \ref{fig:delta}. }
\label{fig:polylines}
\end{figure}

\begin{figure}[t]
\centering
\includegraphics[width=185pt]{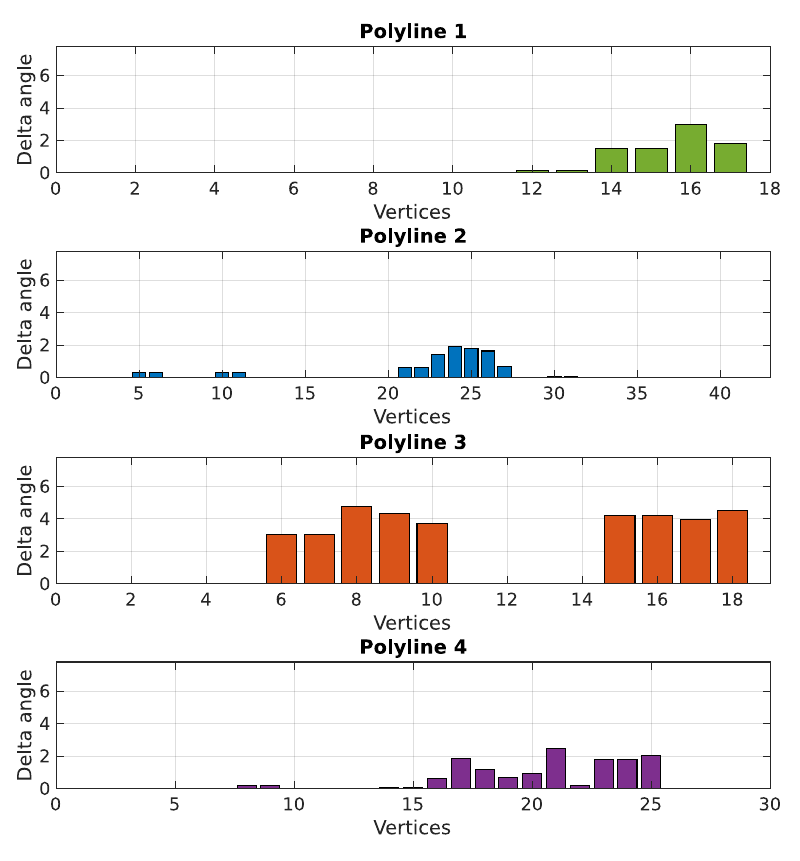}
\caption{Example of 1D signals $\mathbf{\Delta}_h$ derived from polylines represented in Fig. \ref{fig:polylines} with different colors.}
\label{fig:delta}
\end{figure}

%%%%%%%%%%%%%%%%%%%%%%%%%%%%%%%%%%%%%%%%%%%%%%%%%%%%%%%%%%%%%%%%%%%%%%%%%%%%%%%%

\section{GRAPH MODEL DEFINITION}
\label{sec:optimization}

We assume that we have a prior localization (called odometry from now on) in world frame coordinates. Such localization could come from GNSS fusion with odometry systems or SLAM approaches, among others. We define the odometry trajectory as $\hat{\mathbf{X}} = ( \hat{\mathbf{x}}_1, ... , \hat{\mathbf{x}}_N )$, where $\hat{\mathbf{x}}_i \doteq ( \hat{\mathbf{R}}_i,  \hat{\mathbf{t}}_i)$, $\hat{\mathbf{t}}_i \in \mathbb{R}^2$ is the translation, and $\hat{\mathbf{R}}_i \in SO(2)$ is the rotation matrix. We also assume that we have a world’s
representation defined as a set of landmark $\mathcal{L}$. Then, for each $i$-th frame defined by $\hat{\mathbf{X}}$, we observe the landmarks of the environment using on-board sensors. We name these observations as detections $\mathcal{D}_i$ from now on. Using $\mathcal{L}$ and $\mathcal{D}_i$, we perform a data association process, where its result is defined by $\mathbf{A}_i = ((\mathbf{d}_{i_1}, \mathbf{l}_{i_1}), ... , (\mathbf{d}_{i_M}, \mathbf{l}_{i_M}))$. Given these associations, if we express as $\mathbf{X}$ the pose estimated in the optimization process, we can define the residuals between landmarks and detections as follows: 

\begin{equation}
    e_a(\mathbf{X}) = \sum_{i=1}^{N}\sum_{k=1}^{M} \left\lVert \left(\mathbf{R}_i\mathbf{d}_{i_k} + \mathbf{t}_i \right) - \mathbf{l}_{i_k} \right\rVert^2_{\Sigma_{i_k}}
    \label{eq:residuals_lm}
\end{equation}
where $\Sigma_{i_k}$ is the covariance matrix of each detection $\mathbf{d}_{i_k}$ derived in Section \ref{sec:covariance}. The covariance is transformed into an information matrix $\Omega_{i_k} = \Sigma^{-1}_{i_k}$ that weighs the residuals as $e^{\mathsf{T}}_{i_k} \Omega_{i_k} e_{i_k}$. 

Additionally, given the relative transformations from consecutive frames $i$ and $j$ from the odometry trajectory $\hat{\mathbf{X}}$ and the estimated $\mathbf{X}$, we can define the odometry residuals as follows:

\begin{equation}
    e_x(\mathbf{X}) = \sum_{i,j}^N \omega^t_{ij} \left\lVert \mathbf{R}^\mathsf{T}_i\left( \mathbf{t}_j - \mathbf{t}_i \right) - \hat{\mathbf{t}}_{ij} \right\rVert^2_2 + \omega^R_{ij} \left\lVert \mathbf{R}^\mathsf{T}_i\mathbf{R}_j - \hat{\mathbf{R}}_{ij} \right\rVert^2_{F}
    \label{eq:residuals_odom}
\end{equation}
As the odometry information matrix weights each norm, we characterize the noise using the $\omega^t_{ij}$ and $\omega^R_{ij}$ weight variables  that depend on the noise model. The subscript $F$ in the second term indicates the Frobenius norm. Hence, given the residuals, graph-based localization aims to estimate the trajectory $\mathbf{X}^*$ that minimizes the error function:

\begin{equation}
    \mathbf{X}^* = \argmin_{\mathbf{X}} \left( e_a(\mathbf{X}) + e_x(\mathbf{X}) \right).
    \label{eq:optimization}
\end{equation}
This is an optimization problem, and we use a Gauss-Newton Non-linear Least Square (NLS) approach to solve it.

%%%%%%%%%%%%%%%%%%%%%%%%%%%%%%%%%%%%%%%%%%%%%%%%%%%%%%%%%%%%%%%%%%%%%%%%%%%%%%%%
\section{SELF-TUNING DATA ASSOCIATION}
\label{sec:self_tuning}

We consider lane marking the best visible landmarks for aerial imagery and on-board sensors for urban and road scenarios. Then, previously defined $\mathcal{L}$ and $\mathcal{D}_i$ contains these kinds of features. The problem is that given the aliasing risk in the straight roads for lane markings, the associations $\mathbf{A}_i$ usually have outliers. 

In Section \ref{sec:information}, we analyze the detections to quantify how straight the road is by using our DA-LMR lane marking representation \cite{munoz2022lmr}. This is directly related to the information theoretical entropy of the detections. Then,  to mitigate outliers, in Section \ref{sec:association}, we tune the search area of our DC-SAC data association \cite{munoz2022lmr} depending on the local pseudo-entropy of the road.

\subsection{Pseudo-entropy quantification by DA-LMR}
\label{sec:information}
We can think of lane marking as polylines, while our DA-LMR \cite{munoz2022lmr} defines each one as a set of 3D points $\mathbf{P}^{3D} = (\mathbf{p}^{3D}_0, \mathbf{p}^{3D}_1, ..., \mathbf{p}^{3D}_n)$ where

\begin{equation}
    \mathbf{p}^{3D}_l = (x_l, y_l, \Delta\alpha_l w).
    \label{eq:dalmr}
\end{equation}
The third dimension of that representation encodes the differential angle $\Delta\alpha_l$ between adjacent segments in polyline weighed with a configurable parameter $w$. We show in Fig. \ref{fig:polylines} an example of 2D projected polylines for a set of detections $\mathcal{D}_i$. 

To measure how straight the road that the data represents is, for each $h$-th polyline, we extract the delta angle from DA-LMR and describe it as a 1D signal $\mathbf{\Delta}_h = (\Delta\alpha_{h_1}, ... , \Delta\alpha_{h_n})$. In Fig. \ref{fig:delta}, we depicted the 1D signals $\mathbf{\Delta}_h$ from polylines represented in Fig. \ref{fig:polylines}. Then with this representation, we can calculate the information as follows:

\begin{equation}
    S = - \sum_{h=1}^{m} \sum_{l=1}^{n} \Delta\alpha_{h_l} \log (\Delta\alpha_{h_l} + 1).
    \label{eq:entropy}
\end{equation}
The index $h$ refers to entire polylines, while $l$ indexes each element in the polyline. $S$ is not a by-definition entropy as $\mathbf{\Delta}_h$ is not a probabilistic distribution. Nevertheless, regularizing the logarithm by $+1$ term changes its behavior to be similar as entropy, and we can consider $S$ as a pseudo-entropy.

To understand this quantification, we can consider an example of detections on a straight road. In that case, the lane markings are linear, and $\Delta\alpha_{h_l}$ values are close to $0$. Hence, the measure of $S$ is also close to $0$. Then we can consider this case as low informative for the data association process. In contrast, the example depicted in Fig. \ref{fig:polylines} contains polylines with an amount of information concentrated in corners. Then, the value of $S$ can achieve high negative values. Hence, we can consider this as an example of highly informative detections.

\begin{figure}[t]
\centering
\includegraphics[width=200pt]{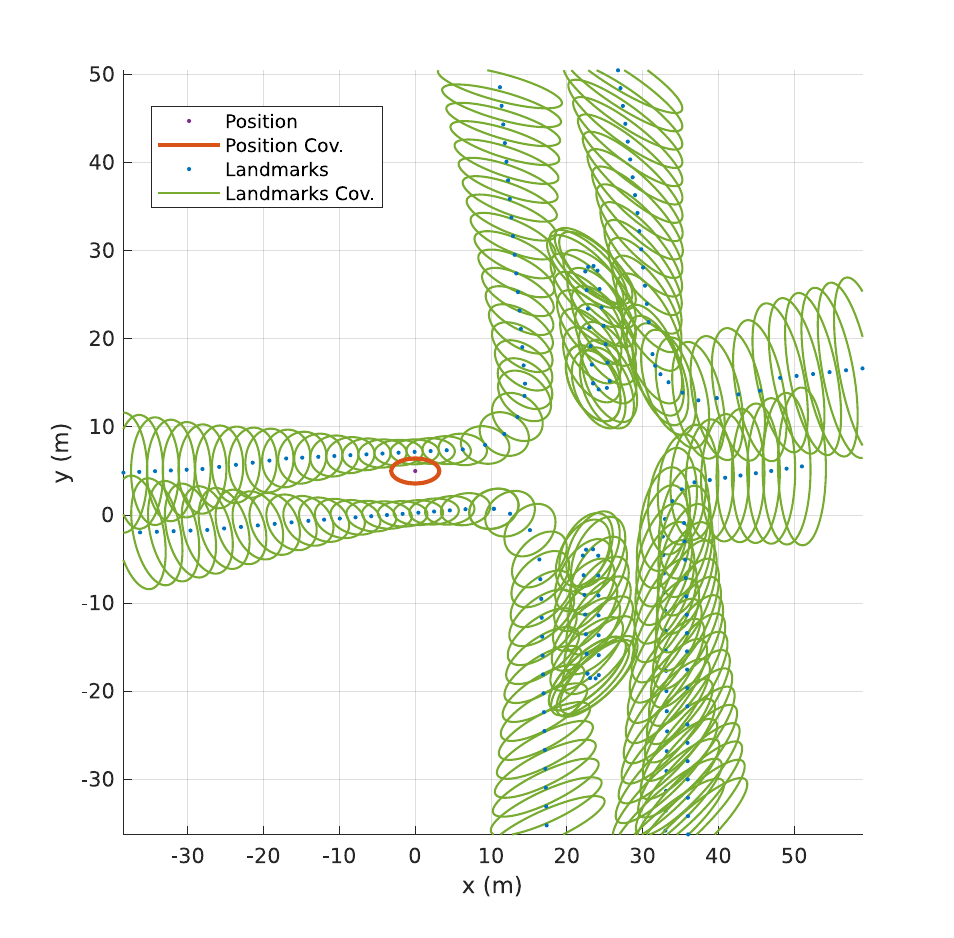}
\caption{Covariance propagation in a set $\mathcal{D}_i$, where the red ellipse indicates the data association covariance $\Sigma_{i}$ and the green ellipses depict the detections covariances propagated $\Sigma_{i_k}$.}
\label{fig:covariance}
\end{figure}

\begin{figure*}[t]
\centering
\includegraphics[width=500pt]{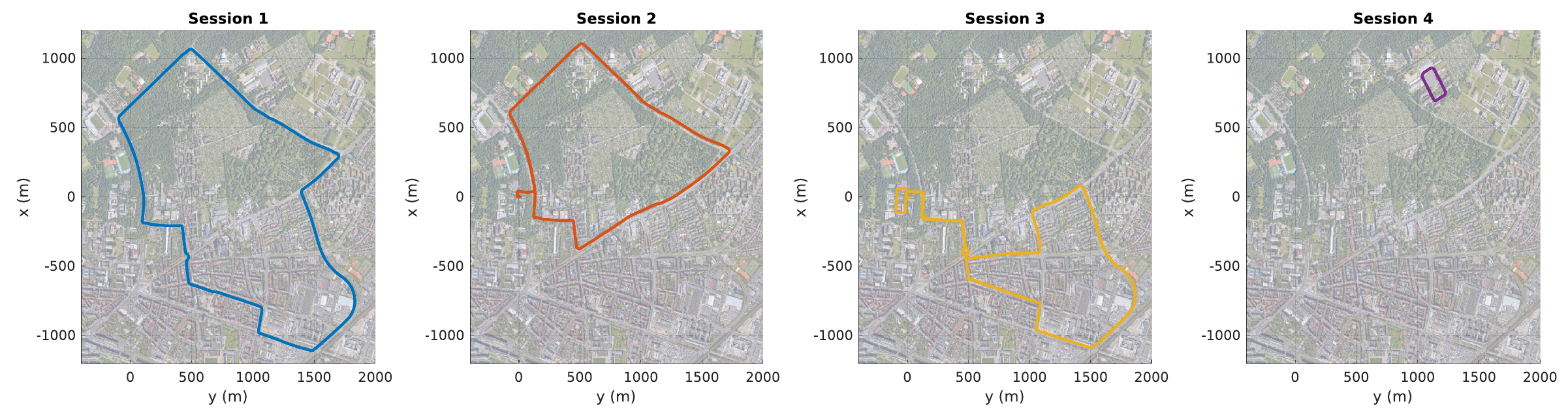}
\caption{The four closed trajectories for evaluation, driving through the city of Karlsruhe (Germany) and its outer roads. In Table \ref{tab:datasets}, we show more details about these trajectories.}
\label{fig:sessions}
\end{figure*}

\subsection{Data association by DC-SAC}
\label{sec:association}
The DC-SAC data association method \cite{munoz2022lmr} randomly samples a pair of points in $\mathcal{D}_i$ and a couple of distance-compatible points in $\mathcal{L}$. Given the samples, we obtain the transform $\Delta \mathbf{T} \in SE(2)$ that minimizes the error between the pairs by solving the Procrustes problem. We repeat this process generating a hypothesis space $\mathcal{H}$. The final association comes from the $\Delta \mathbf{T}^{*} \in \mathcal{H}$ that achieves the minimum error from the inliers between $\mathcal{D}_i (\Delta \mathbf{T}^{*})$ and $\mathcal{L}$.

As explained in \cite{munoz2022lmr}, if the we reduce the sample area size defined by configurable parameters $\Phi = (x_{max}, y_{max}, \theta_{max})$\footnote{In \cite{munoz2022lmr}, the area size configuration is described in text, but only $\theta$ is mathematically defined. Hence, for this work, we derive area size parameters $\Phi$ for the sake of clarity.}, the number of distance-compatible samples in DC-SAC are also reduced. And, consequently, as $\Delta \mathbf{T} \in \mathcal{H}$ come from that samples, the hypothesis space are also reduced. 

We observed that DC-SAC, with proper $\Phi$ configuration, produces excellent results with a highly informative set $\mathcal{D}_i$, as the one shown in Fig. \ref{fig:polylines}. However, in straight roads, it suffers, like others, from an aliasing effect that produces outliers.

It is worth noting that when $\Phi \xrightarrow{} 0$, the space $\mathcal{H}$ only contains a default transform $\Delta \mathbf{T} \triangleq 0$, i.e., DC-SAC will only return the identity transformation. Hence, the associations come from the inliers between $\mathcal{D}_i$ and $\mathcal{L}$, which is equivalent to using classical Nearest Neighbour (NN) radius search with the distance-compatibility threshold as the radius. For straight markings, NN can mitigate that aliasing effect as good as possible given the odometry's information. In contrast, NN does not exploit highly informative environments.

Then, given DC-SAC and NN's complementary behavior, we can self-tune the parameters $\Phi_i$ dynamically for each $\mathcal{D}_i$ by using the pseudo-entropy information as follows: 

\begin{equation}
    \Phi^{tuned}_i = \left\lbrace 
        \begin{array}{ll}
        \Phi, &  S_i \leq S_{min}\\
        \Phi  \frac{S_i}{S_{min}}, & S_i > S_{min}\\
    \end{array}
    \right.
    \label{eq:params}
\end{equation}
$S_{min}$ is a configurable parameter that saturates the limit of the minimum value of pseudo-entropy, thus controlling the ramp for the tuning. Note that always $S_i \leq 0$, then $0 \leq \frac{S_i}{S_{min}} < 1$. $\Phi$ is the maximum area size configured, then $\Phi^{tuned}_i \leq \Phi$. We can see in \eqref{eq:params} that when $S_i \xrightarrow{} 0$ (e.g. straight road), the behavior is close to NN, and when $S_i \xrightarrow{} S_{min}$  (e.g. intersection), the behavior comes to DC-SAC in its maximum sample area size configuration. This area size self-tuning capability mitigates the aliasing effect dramatically and, hence, the outlier risk, as we demonstrate experimentally in Section \ref{sec:evaluation}.

%%%%%%%%%%%%%%%%%%%%%%%%%%%%%%%%%%%%%%%%%%%%%%%%%%%%%%%%%%%%%%%%%%%%%%%%%%%%%%%%

\section{DYNAMIC COVARIANCE ADJUSTMENT}
\label{sec:covariance}

In addition to the already explained self-tuned data association, we implemented another layer of robustness by dynamic covariance adjustment. For each data association process, we estimate a covariance matrix depending on the result of DC-SAC (Section \ref{sec:variance}). Afterward, we propagate that covariance to detections by first-order transformation (Section \ref{sec:propagation}), and we use the inverse as an information matrix in the optimization process.

\subsection{Data association variance}
\label{sec:variance}
We defined in Section \ref{sec:optimization} the odometry trajectory as $\hat{\mathbf{X}} = ( \hat{\mathbf{x}}_1, ... , \hat{\mathbf{x}}_N )$, and in the same way, we can define the estimated trajectory as $\mathbf{X} = (\mathbf{x}_1, ... , \mathbf{x}_N )$. However, in the data association stage, before pose estimation $\mathbf{X}$ at $i$-th sample time, we can propagate the pose by using $\mathbf{x}_{i-1}$ and integrating the last differential of the odometry $\Delta \hat{\mathbf{x}}_{i-1,i}$. Then, we can express the new odometry as $\Bar{\mathbf{x}}_i = \mathbf{x}_{i-1} \Delta \hat{\mathbf{x}}_{i-1,i}$.

DC-SAC data association method is \textit{pose-based}, which means that apart from the associations $\mathbf{A}_i$, the result also contains the relative transform $\Delta \mathbf{T}^{*}_i \in \mathcal{H}_i$ that produces $\mathbf{A}_i$. The transform $\Delta \mathbf{T}^{*}_i$ is relative to previously explained $\Bar{\mathbf{x}}_i$.

If we think in a hypothetical ideal case of localization without errors, the relative transforms $\Bar{\mathbf{T}} = (\Delta \mathbf{T}_1, ... , \Delta \mathbf{T}_N)$ should have $\Delta \mathbf{T}_i \triangleq 0$ in all cases. Hence, in the real case, we can assume that when the evolution of $\Bar{\mathbf{T}}$ is stable, the data association is reliable. In contrast, when $\Bar{\mathbf{T}}$ has a strong variance, the data association results are unreliable. In theory, the stable behavior of $\Delta \mathbf{T}_i$ can also come from the localization to a local minimum. However, we did not have local minimum issues in the experiments because we have the main self-tuning layer described in Section \ref{sec:self_tuning} that can recover the system when a local minimum is reached.

Under the previously-mentioned assumption, we define
the covariance matrix as:

\begin{equation}
    \Sigma_i = \Psi(\Delta \mathbf{T}_{i-W}, ... , \Delta \mathbf{T}_i)
    \label{sec:tf_var}
\end{equation}
where $\Psi$ is a function that calculates, given the origin pose in the sensor frame $\mathbf{p} = (0, 0, 0)$, the pose $\mathbf{p}_i = (x_i^s, y_i^s, \theta_i^s)$ by using $\Delta \mathbf{T}_i$. Afterward, given the set of poses $\mathbf{p}_i$, it infers the covariance matrix $\Sigma_i$. $W$ is a configurable parameter that defines a window to select the last data association results.

\begin{table*}[t]
    \centering
    \caption{Trajectory evaluation by Absolute Trajectory Error (ATE) in meters.}
    \begin{tabular}{c|c c c c c c c c}
          &  &  & \textbf{DC-SAC} & \textbf{Self-Tuning} & \textbf{Self-Tuning} & \textbf{Covariance} & \textbf{Convex}  \\
          
         \textbf{Session} & \textbf{Odometry} & \textbf{NN (static)} & \textbf{(static)} & \textbf{- Cov. Adj.} & \textbf{+ Cov. Adj.} & \textbf{Scaling \cite{agarwal2013robust}} & \textbf{Relaxation \cite{carlone2018convex}}  \\
    \hline
         1 & 2.61 & 4.14 & 11.23 & 0.14 & 0.07 & 0.38 & 0.24 \\
    \hline
         2 & 3.36 & 4.87 & 10.29 & 0.67 & 0.09 & 0.36 & 0.21 \\
    \hline
         3 & 3.35 & 1.67 & 8.54 & 0.25 & 0.06 & 0.29 & 0.27 \\
    \hline
         4 & 2.68 & 0.74 & 2.21 & 0.09 & 0.06 & 0.16 & 0.14 \\
    \hline
    \end{tabular}
    \label{tab:ate_results}
\end{table*}

\begin{figure*}[t]
\centering
\includegraphics[width=450pt]{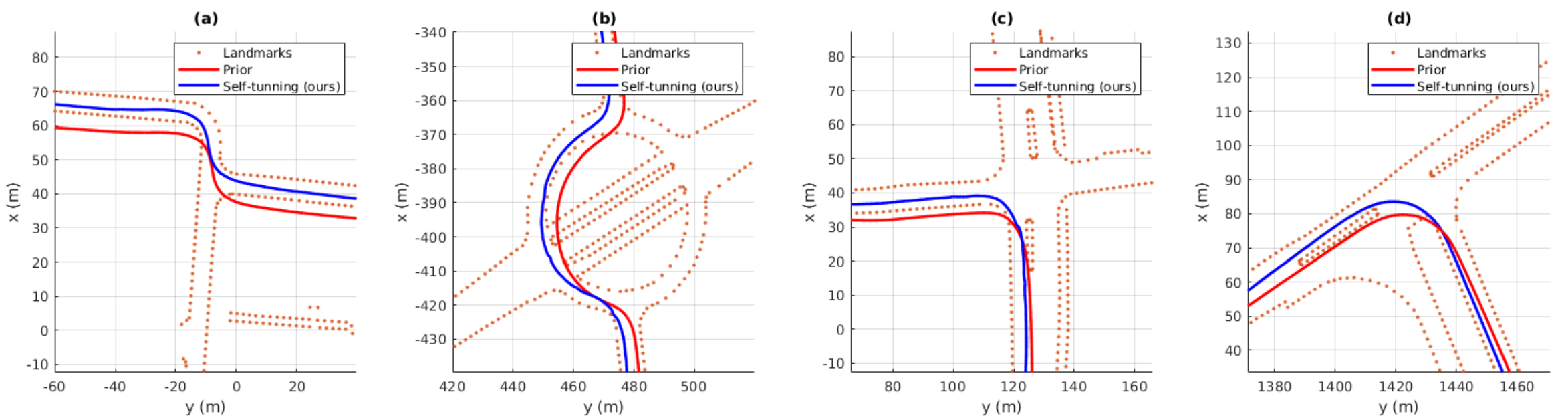}
\caption{Comparison between the odometry trajectories (red lines) and the geo-referenced ones estimated using the proposed self-tuning method (blue lines). The points are the landmarks $\mathcal{L}$ inside a window. }
\label{fig:trajectory}
\end{figure*}

\subsection{Covariance propagation}
\label{sec:propagation}
The covariance matrix $\Sigma_i$ is expressed in the sensor coordinate frame. To make it suitable for constraints \eqref{eq:residuals_lm}, we use first-order transformation to propagate the covariance to detections \cite{roysdon2018technical}, obtaining $\Sigma_{i_k}$. 

We have estimated pose $\mathbf{x}_i \doteq ( \mathbf{R}_i,  \mathbf{t}_i)$, where $\mathbf{t}_i = (x_i, y_i)$. Then, if we express detections as $\mathbf{d}_{i_k} = (x_{i_k}^d, y_{i_k}^d)$ in the sensor coordinate frame, we can transform detections to estimation coordinate frame as:

\begin{equation}
    f_{i_k}^x = x_{i_k}^d \cos \theta_i - y_{i_k}^d \sin \theta_i + x_i
    \label{eq:transform_f1}
\end{equation}
\begin{equation}
    f_{i_k}^y = x_{i_k}^d \sin \theta_i + y_{i_k}^d \cos \theta_i + y_i.
    \label{eq:transform_f2}
\end{equation}
If we compute the first-order derivative, we obtain the Jacobian:

\begin{equation}
    J_{i_k} = 
    \begin{bmatrix}
      \frac{\partial f_{i_k}^x}{x_i} & \frac{\partial f_{i_k}^x}{y_i} & \frac{\partial f_{i_k}^x}{\theta_i} \\
      \frac{\partial f_{i_k}^y}{x_i} & \frac{\partial f_{i_k}^y}{y_i} &    \frac{\partial f_{i_k}^y}{\theta_i}
    \end{bmatrix} =
    \begin{bmatrix}
      1 & 0 & -x_{i_k}^d \sin \theta - y_{i_k}^d \cos \theta \\
      0 & 1 &  x_{i_k}^d \cos \theta + y_{i_k}^d \sin \theta
    \end{bmatrix}
    \label{eq:jacobian}
\end{equation}
Given the Jacobian, we can propagate the covariance to detections as follows:

\begin{equation}
    \Sigma_{i_k} = J_{i_k} \Sigma_{i} J_{i_k}^\mathsf{T}.
    \label{eq:propagation}
\end{equation}
In Fig. \ref{fig:covariance}, we show an example of covariance propagation, where the red ellipse indicates the data association covariance $\Sigma_{i}$ and the green ellipses depict the detections covariances propagated $\Sigma_{i_k}$. 

%%%%%%%%%%%%%%%%%%%%%%%%%%%%%%%%%%%%%%%%%%%%%%%%%%%%%%%%%%%%%%%%%%%%%%%%%%%%%%%%

\section{EVALUATION}
\label{sec:evaluation}

%In this section, we show the results of our evaluation. First (Section \ref{sec:setup}), we comment on the experimental setup. Next (Section \ref{sec:traject}), we assessed the trajectory compared with different configurations and state-of-the-art methods. And finally (Section \ref{sec:mitigation}), we focus on the evaluation of outlier mitigation.

\subsection{Experimental setup}
\label{sec:setup}
The evaluation consists of four closed trajectories driven through the city of Karlsruhe (Germany) and its outer roads (Fig. \ref{fig:sessions} and Table \ref{tab:datasets}). The four sessions are initially referenced with low-cost GNSS measurements using the approach proposed in \cite{sons2018efficient}, and the results are used as a odometry trajectory. Each odometry pose has its corresponding lane markings detections $\mathcal{D}_i$. To perform that process, we used the experimental vehicle \textit{BerthaOne} \cite{tacs2017making}. The car comprises four Velodyne VLP16 LiDARs mounted flat on the roof, three BlackFly PGE-50S5M cameras behind the front and rear windshield, and a Ublox C94-M8P GNSS receiver. To test the proposed approach, we use a hand-made map $\mathcal{L}$ built from geo-referencing aerial images. We manually annotated all the lane markings around the area where we completed the trajectories\footnote{Alternatively, it is possible to use an automatic lane markings extractor~\cite{azimi2018aerial}.}. The lane marking detection (to obtain $\mathcal{D}_i$) is performed using cameras in the stereo configuration by the method described in \cite{poggenhans2015universal}. The lane markings are segmented and then classified using an Artificial Neural Network. Finally, the classified markings are converted into polylines.

\begin{table}[ht!]
    \centering
    \caption{Sessions characteristics description.}
    \begin{tabular}{c c c c}
    \hline
         \textbf{Session} & \textbf{Scan Number} & \textbf{Length (km)} & \textbf{Environment} \\
    \hline
         1 & 5085 & 7.09 & Urban + Rural \\
         2 & 4172 & 5.49 & Urban + Rural \\
         3 & 5105 & 6.63 & Urban \\
         4 & 598 & 0.73 & Rural \\
    \hline
    \end{tabular}
    \label{tab:datasets}
\end{table}

\begin{table*}[t]
    \centering
    \caption{Outlier evaluation by Relative Pose Error (RPE) in translation and rotation.}
    \begin{tabular}{c c|c c c c c c c}
          &  &  & \textbf{DC-SAC} & \textbf{Self-Tuning} & \textbf{Self-Tuning} & \textbf{Covariance} & \textbf{Convex}  \\
          
         \textbf{Session} & \textbf{RPE} & \textbf{NN (static)} & \textbf{(static)} & \textbf{- Cov. Adj.} & \textbf{+ Cov. Adj.} & \textbf{Scaling \cite{agarwal2013robust}} & \textbf{Relaxation \cite{carlone2018convex}}  \\
    \hline
         1 & \textit{trans.} ($m$) & 0.04 & 9.34 & 0.04 & 0.04 & 0.23 & 0.14 \\
           & \textit{rot.} ($\deg$) & 0.07 & 0.14 & 0.06 & 0.06 & 0.08 & 0.07 \\
    \hline
         2 & \textit{trans.} ($m$) & 0.04 & 10.00 & 0.06 & 0.06 & 0.25 & 0.13 \\
           & \textit{rot.} ($\deg$) & 0.07 & 0.13 & 0.12 & 0.12 & 0.16 & 0.14 \\
    \hline
         3 & \textit{trans.} ($m$) & 0.04 & 9.88 & 0.07 & 0.06 & 0.28 & 0.16 \\
           & \textit{rot.} ($\deg$) & 0.08 & 0.17 & 0.11 & 0.11 & 0.15 & 0.12 \\
    \hline
         4 & \textit{trans.} ($m$) & 0.04 & 3.38 & 0.06 & 0.06 & 0.18 & 0.16 \\
           & \textit{rot.} ($\deg$) & 0.08 & 0.13 & 0.11 & 0.09 & 0.14 & 0.10 \\
    \hline
    \end{tabular}
    \label{tab:rpe_results}
\end{table*}

\begin{figure*}[t]
\centering
\includegraphics[width=450pt]{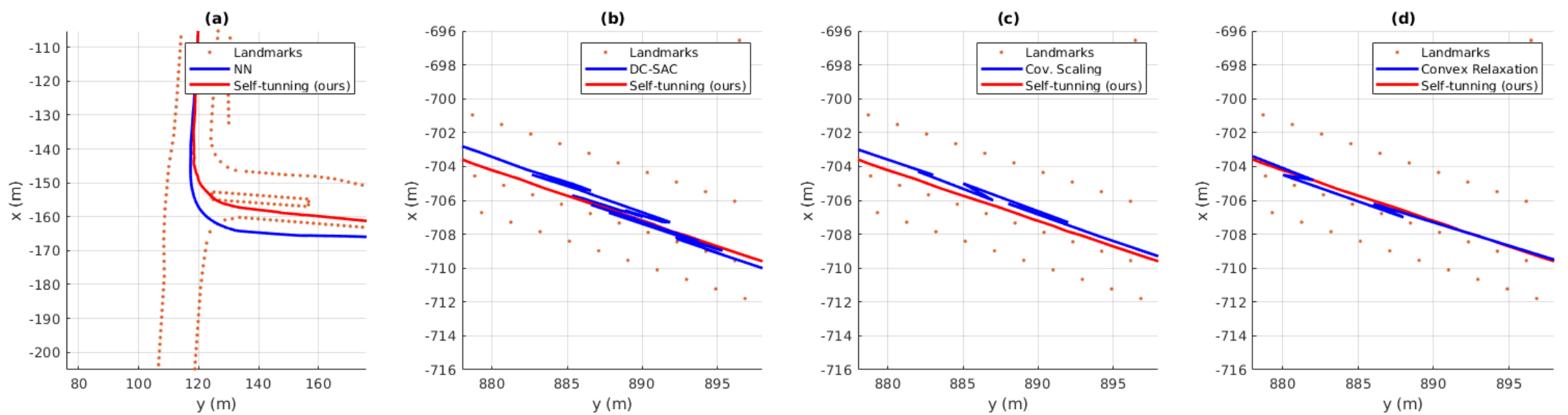}
\caption{Comparison between the estimated trajectories using the proposed \textbf{self-tuning + covariance adjustment} method (red lines) and the compared configurations and SOTA methods (blue lines): (a) NN, (b) DC-SAC, (c) covariance scaling, and (d) convex relaxation.}
\label{fig:outliers}
\end{figure*}

\subsection{Trajectory evaluation}
\label{sec:traject}
Using $\mathcal {L}$ as a reference, we built a hand-made ground truth by the following process: 1) We select an $i$-th sample in the odometry trajectory. 2) We plot detections $\mathcal {D}_i$ generated in the odometry coordinates frame. 3) Next, we transform the detections by hand to coincide with landmarks in  $\mathcal {L}$. 4) Finally, we apply the transform to the $i$-th odometry sample, and the result is our hand-made ground truth. This process results reliably in position but is not precise for orientation. For this reason, we do not evaluate the rotation. This is not an inconvenience in our evaluation because the aliasing problem occurs on a straight road, and produces the error in translation (we demonstrate the low rotation error assumption in the next section).  

Given this ground truth as a reference, we evaluate the odometry trajectory  through the Absolute Trajectory Error (ATE) metric \cite{prokhorov2019measuring}, and compare it with different configurations of our approach: \textbf{NN} (ours with $\Phi = (\SI{0}{\meter}, \SI{0}{\meter}, \SI{0}{\radian})$), \textbf{DC-SAC} (ours with $\Phi = (\SI{5}{\meter}, \SI{5}{\meter}, \SI{0.2}{\radian})$), \textbf{Self-Tuning -Cov. Adj.} (ours without covariance adjustment), and \textbf{Self-Tuning +Cov. Adj.} (our complete approach). Additionally, we compare it with two state-of-the-art (SOTA) methods in outlier mitigation: \textbf{Covariance Scaling} \cite{agarwal2013robust} and \textbf{Convex Relaxation} \cite{carlone2018convex}.

Table \ref{tab:ate_results} shows the results in ATE metric between trajectories estimated from different configurations and methods mentioned and ground truth. Due to ground truth only containing positions, we only calculate ATE for translation. 

Due to its challenging aliasing problems, we can see the most significant error in the DC-SAC configuration. The compared SOTA methods mitigate the aliasing effect but not sufficiently. We can see that we achieve the best results with our self-tuning approach, especially with the combination using the covariance adjustment. We can also see in ATE results that the trajectory is smoothest with covariance adjustment than without it.

The error shown for NN configuration is not produced for the aliasing problem but an offset. If, for example, the vehicle drives through a straight road, the NN configuration accumulates a little error. Then when the intersection is achieved, the NN data association has a lateral error in the new ($\SI{90}{\deg}$) direction of lane markings that can produce an offset, as we will depict in the next section Fig. \ref{fig:outliers} (a).

In Fig. \ref{fig:trajectory}, we show in blue a qualitative examples of trajectories estimated through the proposed self-tuning geo-referenceing. In all cases, the course is located in the center of the lane. In contrast, we show in red the odometry trajectories that always are out of the lanes.

\subsection{Outlier mitigation evaluation}
\label{sec:mitigation}
The odometry used for the evaluation \cite{sons2018efficient} is globally inconsistent (because of the reasons discussed in the Introduction) but locally little noisy (because of the fusion described in \cite{sons2018efficient}). That means the differential information has no considerable errors. Hence, we can use that relative odometry information as a reference to evaluate the effects of outliers by using the alternative Relative Pose Error (RPE) metric \cite{prokhorov2019measuring}.

In Table \ref{tab:rpe_results}, we show RPE results to compare the configurations and SOTA methods named in the previous section. In this case, we can see that the error in the NN configuration is similar to the complete self-tuning approach. This is because, as previously-mentioned, the NN has no aliasing problems, then has no "jumps". However, in Fig. \ref{fig:outliers} (a), we depict the main problem of that configuration, where the red line indicates the trajectory estimated with our complete approach, and the blue line indicates the NN configuration path with the flaw of offset problems that produces localization out of the lane. Note that we show Fig. \ref{fig:outliers} (a) in a different scenario because the weakness of NN is in the intersection parts, whereas the rest of the approaches fail on straight roads. 

\begin{figure}[ht!]
\centering
\includegraphics[width=200pt]{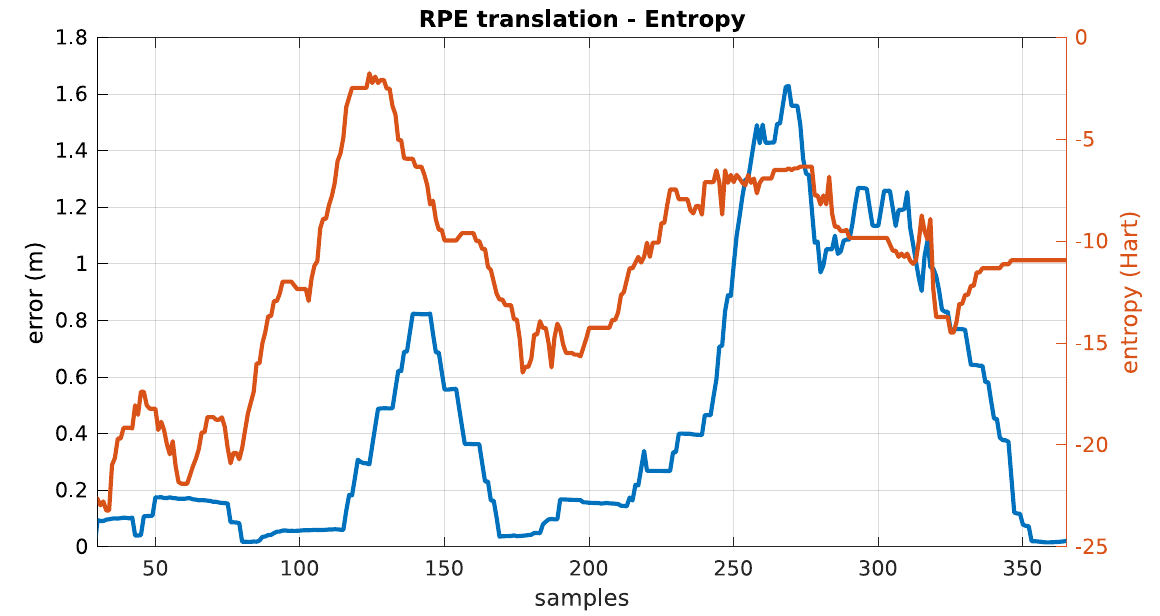}
\caption{The individual pose RPE evolution in an arbitrary trajectory window in contrast with the pseudo-entropy for the same poses. The samples $[0,360]$ represent the frames $[i-360, i]$ in navigation.}
\label{fig:entropy}
\end{figure}

We can see in Table \ref{tab:rpe_results} that the DC-SAC configuration suffers hardly from aliasing problems that produce a considerable number of outliers. In Fig. \ref{fig:outliers} (b), we depict the aliasing effect in a straight road compared to our self-tuning approach. As shown in Table \ref{tab:rpe_results} and Fig. \ref{fig:outliers}, the SOTA methods of covariance scaling and convex relaxation can mitigate the effect of outliers but do not achieve the results of our self-tuning approach for this scenario. The local views in Fig. \ref{fig:outliers} (b-d) show the local effects of aliasing. We can analyze the results in Table \ref{tab:rpe_results} as a global view, assuming that the errors reflect the aliasing that produces “jumps” through the global trajectory.

Our method doesn't eliminate the aliasing effect completely. When the trajectory circulates through a straight road, it sometimes accumulates a few little errors due to the use of an NN-like configuration. Then, when the trajectory enters an intersection, as in contrast with NN, DC-SAC has an area search, it can correct this error producing a little jump. Such little errors are shown in Tables \ref{tab:ate_results} and \ref{tab:rpe_results}.

Finally, to demonstrate our assumption of the relationship between the data pseudo-entropy and the outliers, we show the individual pose RPE evolution in an arbitrary trajectory window for non-tuning DC-SAC configuration in contrast with the pseudo-entropy for the same poses. We can see in Fig. \ref{fig:entropy} that exists a direct correlation.

%%%%%%%%%%%%%%%%%%%%%%%%%%%%%%%%%%%%%%%%%%%%%%%%%%%%%%%%%%%%%%%%%%%%%%%%%%%%%%%%
\section{CONCLUSIONS}
\label{sec:conclusions}

This paper presented a complete geo-referencing pipeline using lane markings as landmarks. To address the outliers problems derived from aliasing, we performed a robust implementation providing self-tuning capabilities to our DC-SAC data association by adapting the search area depending on the pseudo-entropy in the measurements represented by our lane marking representation (DA-LMR). Additionally, to smooth the final result, we adjusted the information matrix for the associated data as a function of the relative transform produced by the DC-SAC. We demonstrated considerable outlier mitigation by the experiments performed in urban and rural scenarios, especially on straight roads without intersections that are usually present in rural areas. The method showed acceptable behavior in those high ambiguity areas for data association compared with other state-of-the-art robust implementations.

%In future work, we plan to extend our approach to new kinds of landmarks, such as walls or even building outlines, and an automatic landmarks extraction from the aerial imagery.

%%%%%%%%%%%%%%%%%%%%%%%%%%%%%%%%%%%%%%%%%%%%%%%%%%%%%%%%%%%%%%%%%%%%%%%%%%%%%%%%

\bibliography{references.bib}{}
\bibliographystyle{IEEEtran}

\end{document}